%% file: main.tex
\documentclass[runningheads]{llncs}

\usepackage{eccv}

\usepackage{eccvabbrv}

\usepackage{graphicx}
\usepackage{booktabs}

\usepackage[accsupp]{axessibility}  % Improves PDF readability for those with disabilities.

\usepackage{hyperref}

\usepackage{orcidlink}
\usepackage{colortbl} % 加载colortbl包
\definecolor{mygray}{gray}{0.9}

\usepackage{multirow}
\usepackage{multicol} 

\usepackage{tabularx}
\usepackage{adjustbox}
\usepackage{wrapfig}
\usepackage{caption}
\usepackage{marvosym}

\begin{document}

\title{Semi-Supervised Video Desnowing Network via Temporal Decoupling Experts and Distribution-Driven Contrastive Regularization} 

\titlerunning{Semi-Supervised Video Desnowing Network}

\author{Hongtao Wu\inst{1}\orcidlink{0009-0007-4863-5119} \and
Yijun Yang\inst{1}\orcidlink{0000-0003-4083-5144} \and
Angelica I. Aviles-Rivero\inst{3}\orcidlink{0000-0002-8878-0325} 
\and
Jingjing Ren\inst{1}\orcidlink{0000-0003-1631-9617}
\and
Sixiang Chen\inst{1}\orcidlink{0009-0003-6837-886X}
\and
Haoyu Chen\inst{1}\orcidlink{0000-0001-7618-9733}
\and
Lei Zhu\inst{1,2}\orcidlink{0000-0003-3871-663X}\textsuperscript{\Letter}
}

\authorrunning{Wu et al.}

\institute{The Hong Kong University of Science and Technology (Guangzhou), China \and
The Hong Kong University of Science and Technology,  Hong Kong SAR, China \\
\email{leizhu@ust.hk}\\
\and
University of Cambridge, UK}

\maketitle

% \begin{abstract}
%   The abstract should summarize the contents of the paper. 
%   LNCS guidelines indicate it should be at least 70 and at most 150 words.
%   Please include keywords as in the example below. 
%   This is required for papers in LNCS proceedings.
%   \keywords{First keyword \and Second keyword \and Third keyword}
% \end{abstract}

\input{sec/0_abstract}    
\input{sec/1_intro}

\input{sec/2_Related}
\input{sec/4_method}
\input{sec/5_Experiments}

\input{sec/6_Conclusion}

\clearpage  % TODO REVIEW/FINAL: This \clearpage needs to be removed from both review and camera-ready versions.

\section*{Acknowledgements}
This work is supported by the Guangzhou-HKUST(GZ) Joint Funding Program (No. 2023A03J0671), the National Natural Science Foundation of China (Grant No. 61902275), the Guangzhou Industrial Information and Intelligent Key Laboratory Project (No. 2024A03J0628), and Guangzhou-HKUST(GZ) Joint Funding Program (No. 2024A03J0618). This work is also supported by HKUST(GZ) College of Future Technology Red Bird MPhil Program and Yongjiang Technology (Ningbo) Co., Ltd.

%Please insert your acknowledgments here.
%Excluding references and acknowledgments, your paper is no longer than 14 pages.

\bibliographystyle{splncs04}
\bibliography{main}
\end{document}

%% file: sec/0_abstract.tex
\begin{abstract}

Snow degradations present formidable challenges to the advancement of computer vision tasks by the undesirable corruption in outdoor scenarios.
While current deep learning-based desnowing approaches achieve success on synthetic benchmark datasets, they struggle to restore out-of-distribution real-world snowy videos due to the deficiency of paired real-world training data.
To address this bottleneck, we devise a new paradigm for video desnowing in a semi-supervised spirit to involve unlabeled real data for the generalizable snow removal. 
Specifically, we construct a real-world dataset with 85 snowy videos, and then present a Semi-supervised Video Desnowing Network (SemiVDN) equipped by a novel Distribution-driven Contrastive Regularization. The elaborated contrastive regularization mitigates the distribution gap between the synthetic and real data, and consequently maintains the desired snow-invariant background details. Furthermore, based on the atmospheric scattering model, we introduce a Prior-guided Temporal Decoupling Experts module to decompose the physical components that make up a snowy video in a frame-correlated manner.
We evaluate our SemiVDN on benchmark datasets and the collected real snowy data. The experimental results demonstrate the superiority of our approach against state-of-the-art image- and video-level desnowing methods. Our code and the dataset are available at \href{https://github.com/TonyHongtaoWu/SemiVDN}{https://github.com/TonyHongtaoWu/SemiVDN}.

 \keywords{Video desnowing \and Semi-supervised learning \and Mixture of experts \and Contrastive learning}
\end{abstract}

%% file: sec/1_intro.tex
\section{Introduction}
\label{sec:Introduction}

\begin{figure}[t]
% \centering
    \includegraphics[width=1\textwidth]{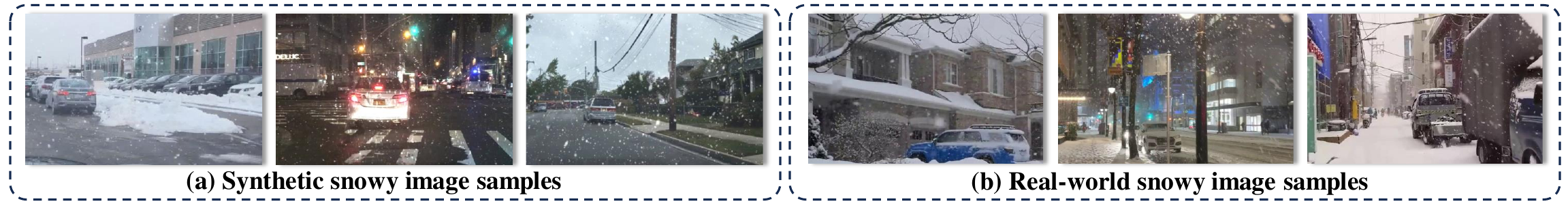}
    % \vspace{-5mm}
    \caption{The distribution shift of the synthetic snow and real snow. }
    \label{fig:intro}
    % \vspace{-3mm}
\end{figure}

\begin{figure*}[t]
\centering
    \includegraphics[width=1\textwidth]{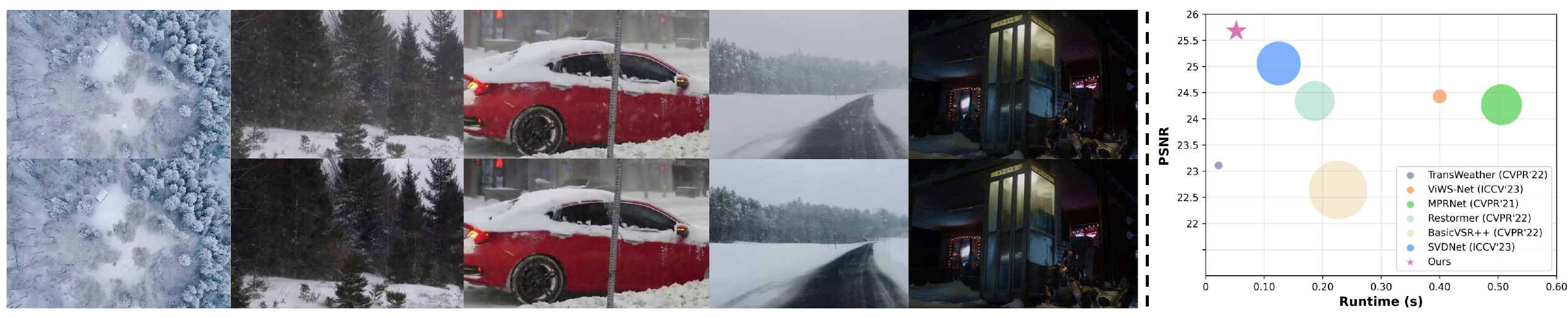}
% \vspace{-5mm}
    \caption{\noindent{\bfseries Left sub-figure:} The proposed semi-supervised method trained using synthetic and real videos yields favorable results on snowy  video samples captured in various scenarios, including forests, country roads and movies. \noindent{\bfseries Right sub-figure:} Trade-off between PSNR performance v.s Runtime and GFLOPs on RVSD dataset~\cite{chen2023snow}. }
    \label{fig:show}
    % \vspace{-5mm}
\end{figure*}

Snow, one kind of adverse weather, frequently appears in outdoor videos. The degradation effects caused by snow particles and streaks severely impair the visibility of video frames and subsequently hinder the advanced performance of video processing algorithms in autonomous systems. Consequently, as an ill-posed inverse problem, a large quantity of desnowing methods are designed to prevent the perturbation of snow not only in single-image but also in videos. 
Early image snow removal methods tend to perform snow removal based on physical priors~\cite{bossu2011rain, pei2014removing, zheng2013single,ren2017video}. Subsequently, deep neural networks like Convolutional Neural Networks (CNNs) and Transformers~\cite{liu2018desnownet, chen2020jstasr, chen2021all,zhang2021deep,li2021online,chen2022snowformer,wang2023masked, xu2023map} are introduced to remove the snow more sophisticatedly. 
According to the atmospheric scattering model~\cite{mccartney1976optics}, Chen \etal~\cite{chen2023snow} constructed a video desnowing benchmark and the 
degradation caused by the snow could be formulated by:
\begin{equation}
\boldsymbol{I}_{snow}\left(\boldsymbol{x}\right)=\boldsymbol{J}\left(\boldsymbol{x}\right)\boldsymbol{T}\left(\boldsymbol{x}\right) + \boldsymbol{A}\left(\boldsymbol{x}\right)\left(1-\boldsymbol{T}\left(\boldsymbol{x}\right)\right)+\boldsymbol{S}\left(\boldsymbol{x}\right) \thinspace,
\label{eq:prior}
\end{equation}
where $\boldsymbol{I}_{snow}$ denotes the video deteriorated by snow, $\boldsymbol{J}$ is the corresponding clean video, $\boldsymbol{T}$ is the transmission map, $\boldsymbol{A}$ is the atmospheric light and $\boldsymbol{S}$ is the snow map.
They also proposed the first network SVDNet to leverage temporal redundancy for snow removal task.
Though such method has achieved success on synthetic benchmarks, better results in recovering real-world snowy videos are highly desired for deployment in real applications.

Unfortunately, as shown in Figure~\ref{fig:intro}, due to the distribution shift between synthetic data and real-world data, it inevitably happens that the existing desnowing methods are impeded by unrealistic training data and fail to handle the real snow with unpredictable shapes and motions. 
More importantly, it's impractical to train these models with plenty of paired real-world data because variable weather conditions, object and camera position make it extremely complicated to align the videos from the real scene.
To address the aforementioned issues, it's a natural practice to consider unlabeled real-world data into the training stage in a semi-supervised fashion. 

In this work, we collect 85 unpaired real-world snowy videos for the training of the proposed model. We use the mixed set composed of synthetic and real data under the Mean-Teacher architecture to enhance its generalizable capability across various real scenarios. 
Specifically, we introduce a Distribution-driven Contrastive Regularization to prevent the deep model from the perturbation of diverse snow shape and motion from synthetic and real data. 
To obtain ultra-positive samples, we utilize the GMM likelihood to capture the synthetic snow most similar to the real counterpart by approximating the distribution of real snow components. 
We maintain and highlight the snow-invariant information by replacing the snow-specific counterpart in ultra-positive samples and contrarily replacing the background in negative samples.

Furthermore, though SVDNet \cite{chen2023snow} manipulates the physical prior of Eq.~\ref{eq:prior}, they lack the beneficial guidance to present an explicit decoupling on each component. To this end, we improve the vanilla transformer block to a physics-based counterpart by introducing temporal decoupling experts. These experts explicitly attend to different compositions of the degradation based on the physical formula, which provides the decomposed feature for the subsequent recovery. We also introduce temporal decomposition router aggregating complementary information within videos to explore the correlations between consecutive frames.

Our contributions can be summarized as:
\begin{itemize}
    \item To the best of our knowledge, we present the first semi-supervised video desnowing framework named SemiVDN, which explicitly explores the beneficial knowledge from unlabeled data to improve the generalization capability of the deep model.
    \item We introduce a Prior-guided Temporal Decoupling Experts module to explicitly decompose the physical components considering inter-frame coherence for better snow removal.
    \item We also design a Distribution-driven Contrastive Regularization to mitigate the appearance difference between the synthetic and real data, and consequently maintain the desired snow-invariant information. 
    \item Extensive experimental results on both synthesized videos and real-world snowy videos demonstrate that our network significantly outperforms other state-of-the-art snow removal methods. More importantly, it  surpasses previous methods in trade-off and performance substantially and has a better generalization ability to benefit real-world applications as shown in Figure~\ref{fig:show}.
\end{itemize}

%% file: sec/2_Related.tex
\section{Related Work}

%-------------------------------------------------------------------------
\subsection{Snow Removal Methods}

Prior to the advent of deep learning~\cite{fang2024spiking, ren2024rethinking,yang2023diffmic,yang2024vivim,xing2024segmamba,fan2024driving, wang2023dynamic}, snow removal techniques \cite{bossu2011rain, pei2014removing, zheng2013single,ren2017video} predominantly relied on physics-based priors to address the snow removal challenge. In recent years, deep-learning-based methods\cite{liu2018desnownet, chen2020jstasr, chen2021all,zhang2021deep,li2021online, chen2023cplformer, Yang_2024_CVPR, wu2023mask, yang2023video, chen2023uncertainty} have achieved impressive results for snow removal. JSTASR \cite{chen2020jstasr} proposed a snow removal algorithm that can jointly classify snow particles and remove the snow with different transparency. HDCW-Net \cite{chen2021all} utilized a hierarchical decomposition paradigm, incorporating dual-tree wavelet transform and wavelet loss. DDMSNet \cite{zhang2021deep} exploited semantic and depth priors for image snow removal. Li \etal \cite{li2021online} proposed an online multi-scale convolutional sparse coding model for online snow removal. Previous research has primarily focused on single-image snow removal techniques, neglecting the complexities of video sequences under snowfall conditions. SVDNet \cite{chen2023snow} presents a video desnowing network with a snow-aware temporal aggregation module by integrating optical flow and snow features to guide the detection and removal of remaining snow within the video sequence. 
However, the aforementioned methods were only trained on synthetic data and may degenerate when deployed on real images caused by the distribution shift. 

%-------------------------------------------------------------------------
\subsection{Semi-Supervised Learning}
In recent years, semi-supervised learning\cite{wang2024advancing, zhu2005semi, wang2023advancing, wang2024dual} has played an increasingly important role in tackling computer vision problems. Researchers introduced semi-supervised methods to learn real data patterns in image restoration tasks. Wei \etal~\cite{wei2019semi} developed a semi-supervised image deraining model using a likelihood term from a parameterized distribution designed for residuals. S2VD \cite{yue2021semi} proposed a semi-supervised video deraining model with a dynamical rain generator.
Recently, many semi-supervised methods have been developed, such as Mean-Teacher \cite{tarvainen2017mean} and MixMatch \cite{berthelot2019mixmatch}.
Among them, the Mean-Teacher method often manipulates consistency regularization based on the high-quality pseudo-labels obtained by an exponential moving average network, which triggers its applications to vision tasks such as semantic segmentation \cite{chen2020multi} and image restoration \cite{liu2021synthetic, wang2022semi}.
DMT-Net \cite{liu2021synthetic} utilized a disentangled-consistency network ensuring consistency between coarse predictions and refinements of real data for image dahazing. 
Wang \etal~\cite{wang2022semi} leveraged a student-teacher framework via knowledge transfer for image super-resolution.
To the best of our knowledge, semi-supervised learning has not yet been explored in the video snow removal task.

%-------------------------------------------------------------------------
\subsection{Mixture of Experts}

Motivated by various successful cases of the Mixture of Experts (MoE) ~\cite{jacobs1991adaptive} in recent advances natural language processing (NLP) tasks~\cite{fedus2022switch,shazeer2017outrageously,shen2023flan, wang2023video}, especially the large language model (LLM), sparse MoE have been popular in high-level vision tasks~\cite{ahmed2016network,gross2017hard,wang2020deep, abbas2020biased, fan2022m3vit, puigcerver2023sparse, yuan2024auformer} due to scaling up module capacity without sacrificing computational cost. Specifically, MoE involves a set of expert networks and a gating network, where gating scores from the gating network adjust the expert networks' outputs. In the community of low-level vision, DRSformer~\cite{chen2023learning} introduces a mixture of experts feature compensators to perform a collaborative refinement of data and content sparsity for image deraining.
Rather than adopting a sparse and discrete router, all the weights in our Temporal Decoupling Experts are continuously considered, while the physics-driven formula implicitly trains the corresponding experts.

%-------------------------------------------------------------------------
\subsection{Contrastive Learning}
Contrastive learning, an efficient self-supervised learning method \cite{oord2018representation, he2020momentum, chen2020simple,yang2023mammodg, xing2024hybrid}, aims to bring anchors closer to positive samples while distancing them from negative samples in the representation space. 
Some works have explored such the paradigm in low-level vision tasks~\cite{wang2021towards,wu2021contrastive}. 
They adopted the original images as positive instances and the degraded images as negative instances, which were subsequently projected into the feature space via VGG \cite{simonyan2014very} for contrastive learning. Semi-UIR \cite{huang2023contrastive} constructed a reliable bank to get the highest image quality samples as pseudo ground truth, which applied contrastive learning on unlabeled data. SVDNet \cite{chen2023snow} pioneered the application of contrastive learning in the desnowing task, based on the observation that distinct videos exhibit unique snow features, whereas identical videos maintain consistent snow features.

%% file: sec/4_method.tex
\section{Methods}
\label{sec:intro}

\begin{figure*}[t]
    \includegraphics[width=\textwidth]{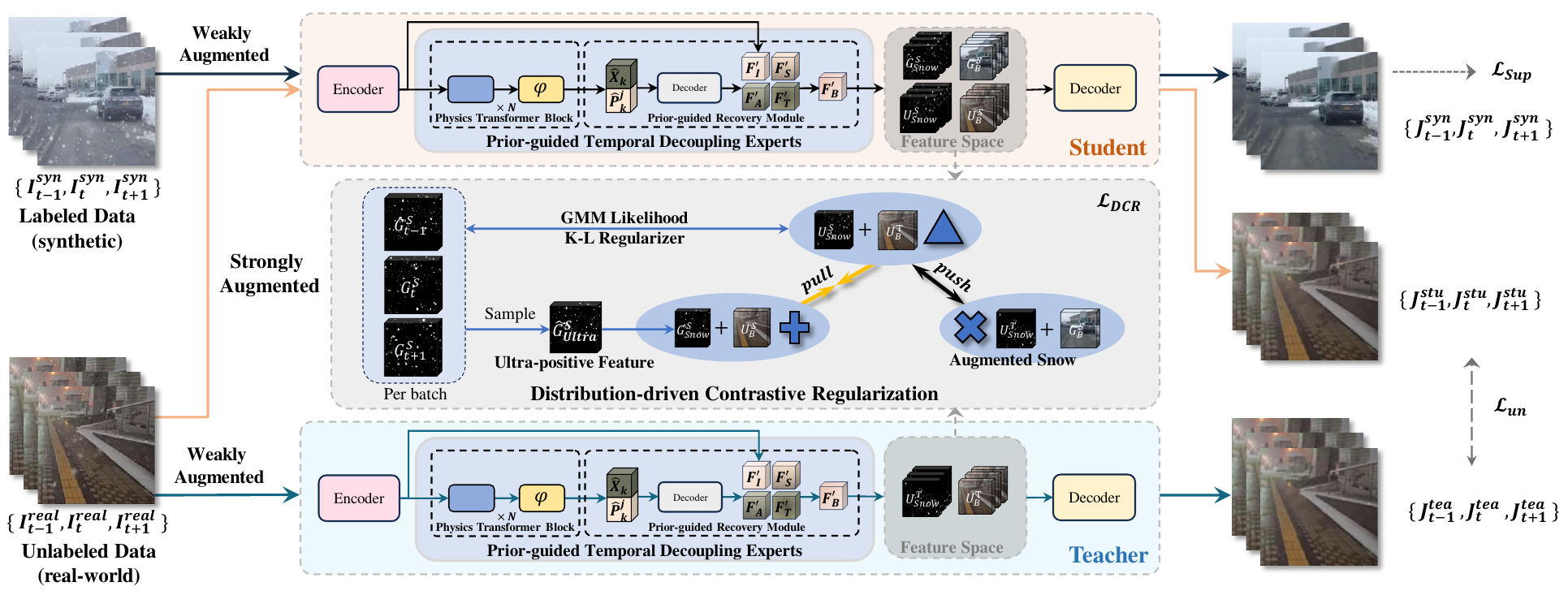}
    % \vspace{-5mm}
    \caption{
    \noindent{\bfseries The schematic illustration of our Semi-Supervised Video Desnowing Network (SemiVDN). } SemiVDN is based on the mean teacher scheme with a student model and a teacher
model. 
% %
We first develop a Prior-guided Temporal Decoupling Experts (see Fig.~\ref{fig:method2}) to decompose the physical components that make up a snow video in a temporal spirit. After that, we compute supervised losses for labeled data and unsupervised losses for unlabeled data. 
Based on the decomposed component features ($F^{'}_{B}$ 
 and $F^{'}_{S}$ ) in representation space, we develop a Distribution-driven Contrastive Regularization to highlight the snow-invariant information by replacing the snow-specific feature in ultra-positive samples and replacing the background in negative samples. }
    \label{fig:method}
    % \vspace{-3mm}
\end{figure*}

%-------------------------------------------------------------------------
\subsection{Network Architecture}

Figure~\ref{fig:method} illustrates the overall framework of the SemiVDN for video snow removal, which is constructed based on the Mean-Teacher fashion~\cite{tarvainen2017mean}. 
Specifically, we develop a video desnowing network (VDN) consisting of an encoder, a novel Prior-guided Temporal Decoupling Experts module, and a decoder. 
Given a snowy video sequence $\{\boldsymbol{I}_{k} \in \mathbb{R}^{{3\times h \times w}} \mid k \in[0, N_f)\}$, we adopt a universal backbone ConvNeXt~\cite{liu2022convnet} as the encoder to extract the feature maps of frames. 
The extracted feature is further fed into our Prior-guided Temporal Decoupling Experts module, which aims to obtain the physical prior components that make up snow videos and remove undesired snow based on Eq.~\ref{eq:prior}. After that, we feed the output desnowed background feature into a decoder to get the final prediction $\{\boldsymbol{J}_{k} \in \mathbb{R}^{{3\times h \times w}} \mid k \in[0, N_f)\}$. 
During the supervised stage, the labeled data is fed into the student network, and pixel-wise supervised loss is computed between the restored and clean frames.
In the semi-supervised stage, we feed the unlabeled data into the student and teacher network, and compute the pixel-wise consistency loss, perceptual contrastive loss and prior losses to regularize the student network.
Moreover, we also exploit Distribution-driven Contrastive Regularization Loss to prevent the model from the negative effects of the distribution gap between synthetic and real data. In the testing stage, we utilize the student network to predict the desnowed results from the input frames.

\begin{figure}[t]
\centering
    \includegraphics[width=0.65\textwidth]{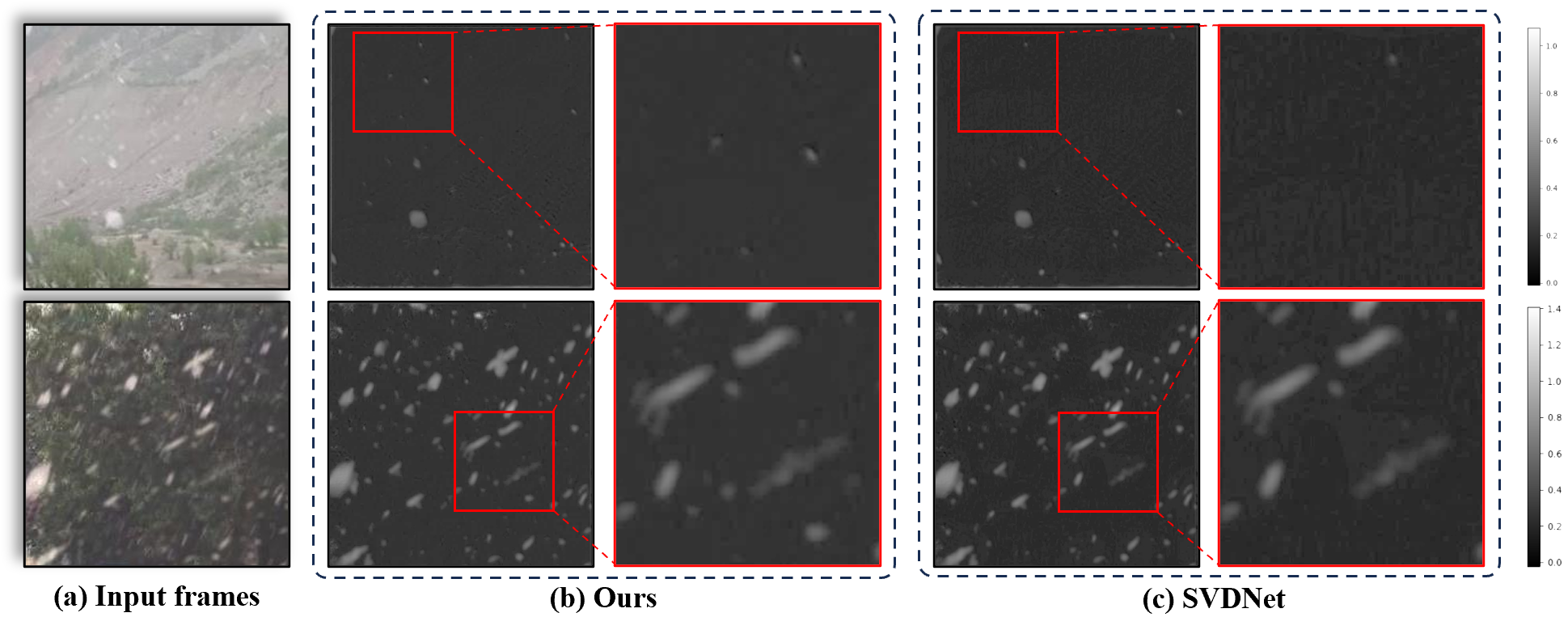}
    % \vspace{-3mm}
    \caption{Comparison of the snow layer decomposition results. It indicates our method can decouple more accurate and clean snow layers without background interference.
}
\label{fig:snow}
% \centering
    % \vspace{-5mm}
\end{figure}
%-------------------------------------------------------------------------

\subsection{Prior-Guided Temporal Decoupling Experts}
Previous works \cite{chen2020jstasr, chen2023snow} tended to remove snow by mimicking its physical model in image and video. 
For example, SVDNet~\cite{chen2023snow} decouples the fused feature and derives several physical features by the convolutional layers based on the formula. 
As shown in Figure~\ref{fig:snow}, this approach frequently fails to capture clean decoupled features (i.e. snow lay feature). To enhance the decoupling ability of the network, we first define a physics transformer block with a set of experts that decomposes the backbone feature into several representative physics-specific features.

\begin{figure*}[t]
\centering
    \includegraphics[width=1\textwidth]{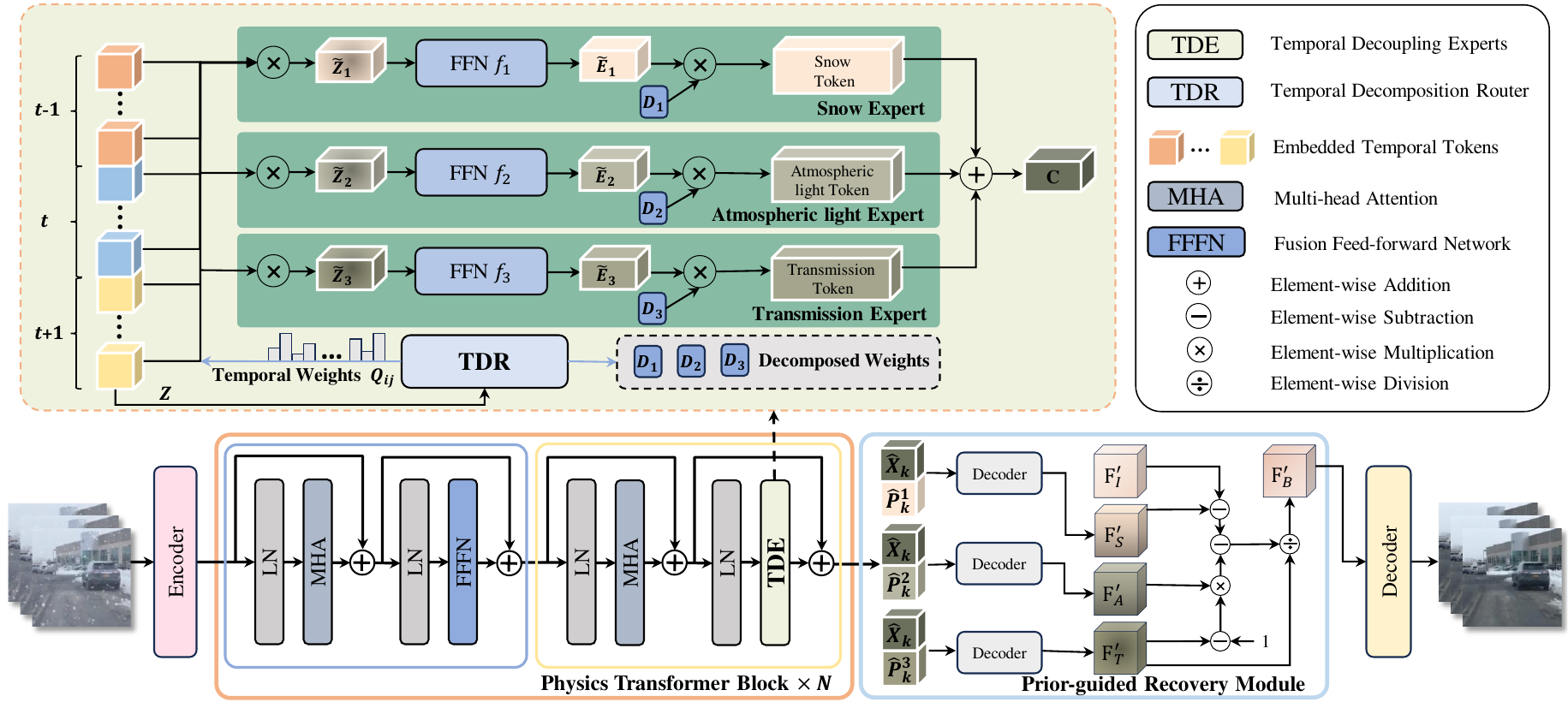}
    % \vspace{-5mm}
    \caption{
\noindent{\bfseries Illustration of the proposed Prior-guided Temporal Decoupling Experts framework. }
Given an input snowy sequence, Physics Transformer Block (PTB) accepts encoded features as input and employs Temporal Decoupling Experts module to generate physics-specific components (i.e. S, A and T) for recovery. 
Specifically, we utilize the Temporal Decomposition Router to compute the temporal weights $\mathbf{Q}_{i j}$ from the temporal dimension, which are subsequently employed to compute a linear combination of all input temporal tokens and $\mathbf{Q}_{i j}$.
Then each expert (an MLP in this work) processes its temporal adaptive tokens to obtain the corresponding output component tokens $\tilde{\mathbf{E}}_{j}$. 
Finally, we employ the decomposed weights from Temporal Decomposition Router to convexly combine all the component tokens.
The output combined features ${\hat{X}}_{k}$ and physics-specific features ${\hat{P}}^{j}_{k}$ are subsequently input into the Prior-guided Recovery Module and the decoder to generate the ultimate desnowed results. 
}
\label{fig:method2}
% \vspace{-5mm}
\end{figure*}

\noindent{\bfseries Physics Transformer Block.} Figure~\ref{fig:method2} illustrates the detailed procedure of the proposed Prior-guided Temporal Decoupling Experts module.
The encoder obtains feature maps of frames
$\{\boldsymbol{X}_{k} \in \mathbb{R}^{{h} / 4 \times w / 4 \times c}\mid k \in[0, N_f)\}$, and each $\boldsymbol{X}$ is individually performed overlapped patch embedding. Then all patches are linearly embedded into tokens $\boldsymbol{Y} \in \mathbb{R}^{(N_f \cdot m) \times d}$, where $m$ is the number
of tokens in one frame and $d$ is the token channel. Then, $\boldsymbol{Y}$ is fed into transformer blocks for physics-dependent information separation. In the physics transformer blocks, we first improve the first transformer’s feed-forward network by a fusion feed-forward network \cite{liu2021fuseformer} to enhance feature fusion. 
Inspired by the popular design of sparse MoEs \cite{lepikhin2020gshard, riquelme2021scaling, fedus2022switch, puigcerver2023sparse}, we incorporate our Temporal Decoupling Experts module into the second transformer of the Physics Transformer Block by replacing its MLP blocks. The proposed module consists of specific experts corresponding to diverse physics components, that are snow expert, transmission expert and atmospheric light expert. We denote the input tokens for one sequence by $\boldsymbol{Z} \in \mathbb{R}^{(N_f \cdot m) \times d}$. Every Temporal Decoupling Experts module utilizes a set of $n$ expert functions, specifically denoted as $\left\{f_{j}: \mathbb{R}^{d} \rightarrow \mathbb{R}^{d}\right\}_{j=1}^n$. 
Each expert will process a temporal adaptive token, and each token has a corresponding $d$-dimensional vector of parameters, denoted as $\boldsymbol{\Gamma } \in \mathbb{R}^{d \times n}$. Then we can get the temporal adaptive weight $\mathbf{Q}_{i j}$ based on the temporal dimension $N_f \cdot m$ from the Temporal Decomposition Router as follows:
% , which is the softmax results of temporal dimension of $\mathbf{Z} \cdot \boldsymbol{\Gamma }$:
\begin{equation}
\mathbf{Q}_{i j}=\frac{\exp \left((\mathbf{Z} \boldsymbol{\Gamma })_{i j}\right)}{\sum_{i=1}^{N_f \cdot m} \exp \left((\mathbf{Z} \boldsymbol{\Gamma })_{i j}\right)} \thinspace.
\end{equation}
Consequently, the weighted tokens $\tilde{\mathbf{Z}}\in \mathbb{R}^{n \times d}$ are the convex combination of $N_f \cdot m$ input tokens and $\mathbf{Q}_{i j}$, which contains adaptive temporal information from the input $N_f$ frames:
\begin{equation}
\tilde{\mathbf{Z}}=\mathbf{Q}^{\top} \mathbf{Z} \thinspace.
\end{equation}

Then, the corresponding expert function is applied to each temporal adaptive token $\tilde{\mathbf{Z}}_{j}$ to obtain the output component tokens:
 % \vspace{-2mm}
\begin{equation}
\tilde{\mathbf{E}}_{j}=f_{j}\left(\tilde{\mathbf{Z}}_{j}\right) \thinspace.
\end{equation}
We can perform dynamic decoding of physics-specific features with the obtained representative component tokens and decomposed weights based on the Temporal Decomposition Router.
Finally, the output tokens $\mathbf{C}$ are computed as a convex combination of all output component tokens:
%%%%%%%%%%%%%%%%%%%%%%%%%%
\begin{equation}
\mathbf{D}_{i j}=\frac{\exp \left((\mathbf{Z} \boldsymbol{\Gamma })_{i j}\right)}{\sum_{j=1}^{n} \exp \left((\mathbf{Z} \boldsymbol{\Gamma })_{i j}\right)} \thinspace, \mathbf{C}=\mathbf{D} \tilde{\mathbf{E}} \thinspace,
\end{equation}
where $\mathbf{D}$ is the decomposed weights, \ie, the softmax results across the expert dimension of $\mathbf{Z} \cdot \boldsymbol{\Gamma }$. Sparse MoE algorithms typically have a discrete nature, making them non-differentiable. This can lead to missing information, such as token dropping and expert unbalance when using classical routing mechanisms. In contrast, our Temporal Decoupling Experts employ continuous and differentiable operations. They effectively leverage the information from all temporal tokens and experts, resulting in the extraction of physics-specific features.
Then, the combined tokens sequence $\mathbf{C}$ and each component tokens are temporally transformed to obtain spatial feature maps $\{\boldsymbol{\hat{X}}_k \in \mathbb{R}^{{h} / 4 \times w / 4 \times c}\mid k \in[0, N_f)\}$ and  $\{\boldsymbol{\hat{P}}^{j}_{k} \in \mathbb{R}^{{h} / 4 \times w / 4 \times c}\mid k \in[0, N_f), j \in[1, n]\}$, respectively. After that, each component feature is, respectively, concatenated with the combined counterpart and then fed into their specific decoder to obtain the enhanced component features. The three decoders are composed of three convolutional layers with upsampling, respectively. Finally, these enhanced features are utilized for the final recovery in the subsequent  prior-guided recovery module. 

\noindent{\bfseries Prior-Guided Recovery Module.} We employ the Eq.~\ref{eq:prior} for the simultaneous removal of snow and haze in frames. This model facilitates the decomposition of frames into three distinct components $S$, $A$, and $T$ in the feature space. According to the Eq.~\ref{eq:prior}, the prior-guided recovery process can be formulated as:
\begin{equation}
F^{'}_{B} =\frac{F^{'}_{I}-F^{'}_{S}-(1-F^{'}_{T}) F^{'}_{A}}{F^{'}_{T}+\beta } \thinspace,
\end{equation}
where $F^{'}_{I}\in \mathbb{R}^{N_f \times c \times h \times w}$ is the encoded input feature,  $F^{'}_{S}\in \mathbb{R}^{N_f \times c \times h \times w}$ is the snow feature, $F^{'}_{T}\in \mathbb{R}^{N_f \times c \times h \times w}$ is the transmission feature, $F^{'}_{A}\in \mathbb{R}^{N_f \times 1 \times h \times w}$ is the global atmospheric light feature and the hyper-parameter $\beta$ is set to $10^{-8}$. Finally, we project the output desnowed feature $F^{'}_{B}$ into RGB space with a convolution layer to obtain the snow-free frame $\mathbf{J}$.

\subsection{Semi-Supervised Video Snow Removal}

In order to enhance the generalization ability of our model across real-world data, we introduce semi-supervised learning (SSL) in video snow removal tasks.
SSL enables a learning system to explore complementary information from both labeled synthesized and unlabeled real-world data. 
As illustrated in Figure~\ref{fig:method}, our SSL framework follows the typical setup~\cite{tarvainen2017mean}.
In the training process, 
the student network is updated by minimizing the supervised losses and unsupervised losses, while the teacher network is updated by the exponential moving average (EMA):
\begin{equation}
\theta_{teacher}^{'}=\eta \theta_{teacher}+(1-\eta) \theta_{student} \thinspace,
\end{equation}
where the EMA decay $\eta$ is empirically set as 0.99. With the adoption of this update strategy, the teacher model can promptly aggregate weights that have been acquired in prior training steps.

\noindent{\bfseries Supervised Loss.} To constrain the outputs of the student network, we adopt the Charbonnier loss~\cite{charbonnier1994two} and the perceptual loss~\cite{johnson2016perceptual} to improve the visual quality of the restored results. 
While the L1 Charbonnier loss is commonly utilized, the perceptual loss is to quantify the disparity between the features of the prediction and the ground truth. 
We extract features from the $3$-rd, $8$-th and $15$-th layers of the pretrained VGG-16 \cite{simonyan2014very} to calculate the perceptual loss.
We also introduced the Focal Frequency Loss~\cite{jiang2021focal} to focus the model on the response of different regions in the frequency spectrum to varying artifacts in the image. The overall supervised loss is formulated as:
 \vspace{-2mm}
\begin{equation}
\mathcal{L}_{\text {sup }}=\mathcal{L}_{pixel}+\lambda_{1}\mathcal{L}_{perceptual}+\lambda_{2}\mathcal{L}_{Frequency} \thinspace,
\end{equation}
where $\lambda_{1}$ and $\lambda_{2}$ are the balancing hyper-parameters, empirically set as 0.03 and 10, respectively.

\noindent{\bfseries Unsupervised Loss.} 
Firstly, we adopt the pixel level Charbonnier loss $\mathcal {L}_{pixel}^{'}$ as the unsupervised teacher-student consistency loss to ensure that the two networks generate consistent results. 
Secondly, we follow \cite{wang2021towards,wu2021contrastive} to incorporate the perceptual contrastive loss by constructing the corresponding perceptual features of $J_{t}^{stu}$ and $J_{t}^{tea}$, and $\hat{I}_{t}^{real}$  as the anchor, positive and negative samples, respectively.
Furthermore, inspired by \cite{chen2020jstasr,chen2021all, shao2020domain, li2019semi}, we employ the dark channel prior (DCP)~\cite{he2010single} loss $\mathcal {L}_{DCP}$ and the total variation loss $\mathcal {L}_{TV}$ as the prior losses, which regularize student network to produce results $J_{t}^{stu}$ with similar statistical characteristics of the clear images.
The overall unsupervised loss is expressed as:
 % \vspace{-2mm}
\begin{equation}
\mathcal{L}_{un}=\lambda_{3} \mathcal {L}_{pixel}^{'}(J_{t}^{stu}, J_{t}^{tea}) + \lambda_{4} \mathcal {L}_{cl}(\hat{I}_{t}^{real}, J_{t}^{tea}, J_{t}^{stu}) + \lambda_{5} \mathcal {L}_{DCP}+ \lambda_{6} \mathcal {L}_{TV} \thinspace,
\end{equation}
where $\hat{I}_{t}^{real}$ denotes the strongly augmented
unlabeled degraded video sequence, $J_{t}^{stu}$ and $J_{t}^{tea}$ denote the snow-free result predicted by student model and teacher model, respectively. 
While $\lambda_{3}$, $\lambda_{4}$ ,$\lambda_{5}$ and $\lambda_{6}$ are the balancing hyper-parameters, empirically set as 2, 0.1, 0.1 and 0.5.
To get the description of unsupervised loss functions, please refer to the \textit{Supplementary File} in detail.

Finally, the overall optimization objective of the student network can be formulated as minimizing the following loss:
\begin{equation}
\mathcal{L}_{\text {overall }}=\mathcal{L}_{\text {sup }}+ \mu \mathcal{L}_{un} \thinspace.
\end{equation}
Inspired by \cite{liu2021synthetic, chen2020multi}, we  apply a time-dependent Gaussian warming up function to update the weight $\mu: \mu(r)=\mu_{\max } e^{\left(-5\left(1-r / r_{\max }\right)^{2}\right)}$, where $r$ denotes the current training iteration and $r_{\max }$ is the maximum training iteration.

%-------------------------------------------------------------------------
\subsection{Distribution-Driven Contrastive Regularization}
Consistency regularization struggles to mitigate the distribution gap between synthesized data and real data~\cite{huang2023contrastive}.
Such the domain gap between real and synthetic snow unexpectedly results in numerous incorrect pseudo-labels. The false information will 
 be accumulated by overfitting on such pseudo-labels.
In order to tackle the aforementioned concern, we introduce contrast learning to shift the attention of the network to the recovery of the snow-invariant background details of the snowy video.

As shown in Figure~\ref{fig:method}, base on our physics transformer block, we can get the pair of the background feature ${G}_{B}^{S}$ and snow feature ${G}_{Snow}^{S}$ from labelled data and the background feature ${U}_{B}^{S}$ and snow feature ${U}_{Snow}^{S}$ from unlabelled data in student network.
Additionally, the background feature ${U}_{B}^{T}$ and snow feature ${U}_{Snow}^{T}$ are obtained from unlabeled data in the teacher network.
According to the synthesis formula of snow $\boldsymbol{I}_{S}\left(\boldsymbol{x}\right)=\boldsymbol{J}\left(\boldsymbol{x}\right) +\boldsymbol{S}\left(\boldsymbol{x}\right)$, we recombine the background features and snow layer features from labelled and unlabelled data. 
In order to maintain and highlight the snow-invariant information, we replace the snow-specific counterpart in positive samples and contrarily replace the background in negative samples. Specifically, we set the ${U}_{B}^{S}$ and ${G}_{Snow}^{S}$ as positive samples, ${U}_{B}^{T}$ and ${U}_{Snow}^{S}$ as anchor samples, ${G}_{B}^{S}$ and augmented ${U}_{Snow}^{T}$ as negative samples.

Due to the distribution differences between synthesized and real snow, the snow layers generated by our network often exhibit noticeable differences.
Therefore, our objective is to acquire an ultra-positive sample, representing the synthetic snow layer that closely resembles the characteristics of real snow layers, to serve as a positive sample.
Since the real snow generally contains inherent varied structures due to their different generation states and observed perspectives, they can be represented by a Gaussian Mixture Model (GMM). 
Employing GMM enables precise approximation of the distribution of real snow layers, effectively capturing the diverse modes within the data. The distribution of snow layers can be represented as:
 % \vspace{-3mm}
\begin{equation}
\mathcal{\nu } \sim  \sum_{p=1}^{K} \pi_{p} \cdot \mathcal{N}\left( \mathcal{\nu } \mid \alpha _{p}, \Sigma_{p}\right) \thinspace,
\end{equation}
 % \vspace{-3mm}
where $\pi_{p}$, $\alpha _{p}$, $\Sigma_{p}$ denote the mixture coefficients, Gaussian distribution means and variances, respectively.
By leveraging the GMM, we can effectively quantify the distribution of real and synthetic snow layers. 
This enables us to compute the Kullback-Leibler (KL) Divergence, which serves as a measure of dissimilarity between the Gaussian mixture module obtained from the real snow layer and diverse synthetic snow layers. 
Through a selection process that prioritizes the minimum KL divergence, we are able to identify the ultra-positive synthesized snow layers $\hat{G}_{Ultra}^{S}$ that closely mirrored the distribution characteristics found in real snow layers ${U}_{Snow}^{S}$. 
After constructing the positive and negative samples, we can calculate the distribution-driven contrastive loss as follows:
\begin{equation}
\mathcal {L}_{DCR}=\frac{\mathcal{L}_{\mathrm{L}_ 1}\left({U}_{B}^{T} + {U}_{Snow}^{S}, {U}_{B}^{S} +\hat{G}_{Ultra}^{S}\right)}{\mathcal{L}_{\mathrm{L}_1}\left({U}_{B}^{T} + {U}_{Snow}^{S}, {G}_{B}^{S}+  Aug\left({U}_{Snow}^{T}\right)\right)+\varepsilon} \thinspace,
\end{equation}
where the hyper-parameter $\varepsilon$ is set to $10^{-7}$, $\mathcal{L}_{\mathrm{L}_ 1}(x,y)$ is the $\ell _{1}$-distance loss between $x$ and $y$, and the weight of this loss is set to 0.1. Eventually, we incorporate the  $\mathcal {L}_{DCR}$ into the $L_{un}^{}$ to derive $L_{un}^{'}$.

%% file: sec/5_Experiments.tex
\section{Experiments}
\label{sec:intro}

\subsection{Real-World Snow Video Datasets}

\begin{figure}[t]
% \centering
    \includegraphics[width=\textwidth]{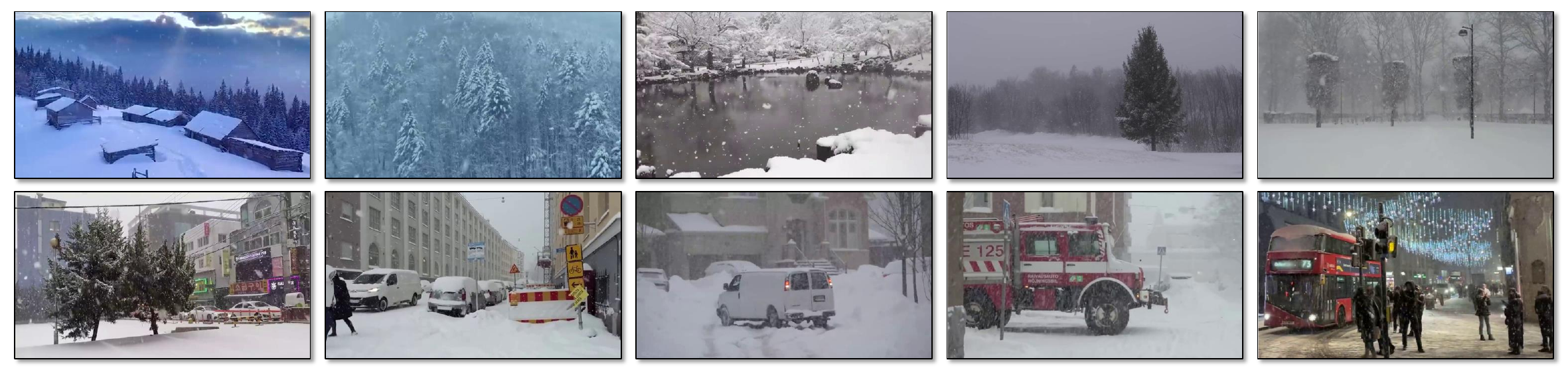}
    % \vspace{-5mm}
    \caption{Samples of the proposed real-world video dataset for video snow removal.}
    \label{fig:buchongreal}
    % \vspace{-3mm}
\end{figure}

In order to tackle the aforementioned issue of lacking suitable video datasets to generalize the performance of desnowing algorithms in real-world snow video, we create the first video snow removal dataset Realsnow85, which is incorporated into the training processing of the semi-supervised video desnowing network.
This dataset serves as a resource for researchers in this field to develop and test novel methods for the removal of snow from video data.
To collect videos for training and testing, we select the snowy video data from the Internet. As shown in Figure~\ref{fig:buchongreal}, we capture different video backgrounds, such as cities, parks, villages and nature. 
In order to enable our model to cope with various snowfall and lighting conditions, we also considered different snowfall levels and lighting scenarios.
In addition, we conduct a comprehensive experiment to evaluate our desnowing network on the Realsnow85 dataset, encompassing 85 videos that exhibit diverse scenes, resolutions, and degradation issues. Among these videos, 60 videos are utilized for training the network, while the other 25 videos are employed for testing and evaluating. 
%%%%%%%%%%%%%
% 
Following \cite{chen2023cplformer, chen2022snowformer}, we use Neural Image Assessment(NIMA)~\cite{talebi2018nima} and Multi-scale Image Quality Transformer(MUSIQ)~\cite{ke2021musiq} as the Non-reference Image Quality Assessment metrics to quantitatively compare the performance of real-world snow degraded video restoration.

\subsection{Implementation Details}
Our network is trained on NVIDIA RTX 4090 GPUs and implemented on the Pytorch platform.
The number of frames per video clip is three.
Each input frame is randomly cropped to a spatial resolution of 256×256. The total number of the training iteration is 300K. We use the AdamW optimizer and the polynomial scheduler. The initial learning rate of our main network is set to $1\times 10^{-4}$ with a batch size of 4.
We set the number of GMM components to be three.

\subsection{Comparison with State-of-the-Art Methods}

\noindent{\bfseries Compared Methods.} To evaluate the effectiveness of the proposed method, we compare it against $15$ state-of-the-art methods, including four snow removal methods~\cite{chen2020jstasr,chen2021all,chen2022snowformer,chen2023snow}, three adverse weather restoration methods~\cite{valanarasu2022transweather,ozdenizci2023restoring,yang2023video}, five fully-supervised restoration methods~\cite{ zamir2021multi,zamir2022restormer,luo2023image,chan2021basicvsr,chan2022basicvsr++}, and three semi-supervised restoration methods~\cite{wu2021contrastive,ye2021closing,yue2021semi}.
For a fair comparison, we implemented these fully-supervised methods by their official codes following \cite{chen2023snow} and retrained them on the RVSD dataset. 
For all compared semi-supervised methods, we follow the same setting of our method to retrain them on a training set, which contains the training set of the RVSD dataset and the training set of our proposed real dataset.
Follow \cite{jinjin2020pipal, gu2020pipal}, we employed the peak signal-to-noise ratio (PSNR), the structural similarity index (SSIM) \cite{wang2004image}, and the learned perceptual image patch similarity (LPIPS) \cite{zhang2018unreasonable} to quantitatively compare the performance between different methods. The average scores of the three metrics are computed for all the frames between the predicted results and the ground truths in the testing set.

\begin{table}[!tbp]
\centering

%\vspace{-3mm}
\caption{Quantitative comparisons between our network and other methods on synthetic datasets and real-world datasets. Bolded and underlined values indicate the best and the second-best performance, respectively.}
% \vspace{-1mm}
\label{tab:resultcompare}
    \resizebox{0.55\columnwidth}{!}{%
\begin{tabular}{c|c|c|ccc|cc} 
\toprule[0.2em]
% \rowcolor{mygray}
\multirow{2}{*}{Method} & \multirow{2}{*}{Type} & \multirow{2}{*}{Venue} & \multicolumn{3}{c|}{Synthetic Datasets} & \multicolumn{2}{c}{Real-world Datasets}  \\

                        &                       &                        & PSNR$\uparrow$  & SSIM$\uparrow$   & LPIPS $\downarrow$                  & NIMA$\uparrow$ & MUSIQ $\uparrow$                        \\ 
\hline
\hline
Input                   &    -                   &  -                      & 18.37 & 0.7792 & 0.3095                 & 4.075    & 48.64                         \\ 
\hline
JSTASR \cite{chen2020jstasr}                  & Image                 & ECCV 2020              & 22.08 & 0.8280  & 0.2336                 & 4.173    & 48.82                         \\
HDCW-Net   \cite{chen2021all}              & Image                 & ICCV 2021              & 22.63 & 0.8592 & 0.2010                  & 4.208   & 47.54                         \\
Snowformer   \cite{chen2022snowformer}           & Image                 & arXiv 2022             & 24.01 & 0.8939 & 0.1219                 & 4.215    & 49.78                         \\
SVDNet    \cite{chen2023snow}               & Video                 & ICCV2023               & \underline{25.06} & \underline{0.9210}  & \underline{0.0842}                 & 4.220    & \underline{50.78}                         \\ 
\hline
TransWeather   \cite{valanarasu2022transweather}          & Image                 & CVPR2022               & 23.11 & 0.8543 & 0.2086                 & 4.182    & 48.06                         \\
WeatherDiffusion  \cite{ozdenizci2023restoring}      & Image                 & TPAMI 2023             & 22.01 & 0.8621 & 0.1539                 & 4.106    & 48.87                         \\
ViWS-Net   \cite{yang2023video}             & Video                 & ICCV2023               & 24.43 & 0.8922 & 0.1142                 & \underline{4.238}    & 50.56                        \\ 
\hline
MPRNet    \cite{zamir2021multi}               & Image                 & CVPR2021               & 24.27 & 0.8960  & 0.1266                 & 4.206    & 50.08                         \\
Restormer    \cite{zamir2022restormer}           & Image                 & CVPR2022               & 24.34 & 0.8929 & 0.1164                 & 4.218    & 50.34                         \\
IR-SDE  \cite{luo2023image}        & Image                 & ICML2023               & 22.71 & 0.8749 & 0.1168                 & 4.099   & 47.45                         \\
IconVSR  \cite{chan2021basicvsr}               & Video                 & CVPR2021               & 22.35 & 0.8482 & 0.2034                 & 4.185    & 49.27                         \\
BasicVSR++  \cite{chan2022basicvsr++}            & Video                 & CVPR2022               & 22.64 & 0.8618 & 0.1868                 & 4.221    & 49.97                         \\ 
\hline
AECR-Net   \cite{wu2021contrastive}             & Image                 & CVPR2021               & 22.95 & 0.8530  & 0.1925                 & 4.188    & 49.81                         \\
JRGR   \cite{ye2021closing}                 & Image                 & ICCV2021               & 23.73 & 0.8729 & 0.1427                 & 4.139    & 48.63                         \\
S2VD    \cite{yue2021semi}                & Video                 & CVPR2021               & 24.02 & 0.8761 & 0.1513                 & 4.156    & 49.61                         \\ 
\hline
\hline
\rowcolor{mygray}\textbf{SemiVDN}                    &  Video                 & -                      & \textbf{25.68} & \textbf{0.9254} & \textbf{0.0785} &  \textbf{4.259}         & \textbf{51.57}                         \\
\bottomrule[0.2em]
\end{tabular}
}
%\vspace{-3mm}
\end{table}

\begin{figure*}[t]
\centering
    \includegraphics[width=0.70\textwidth]{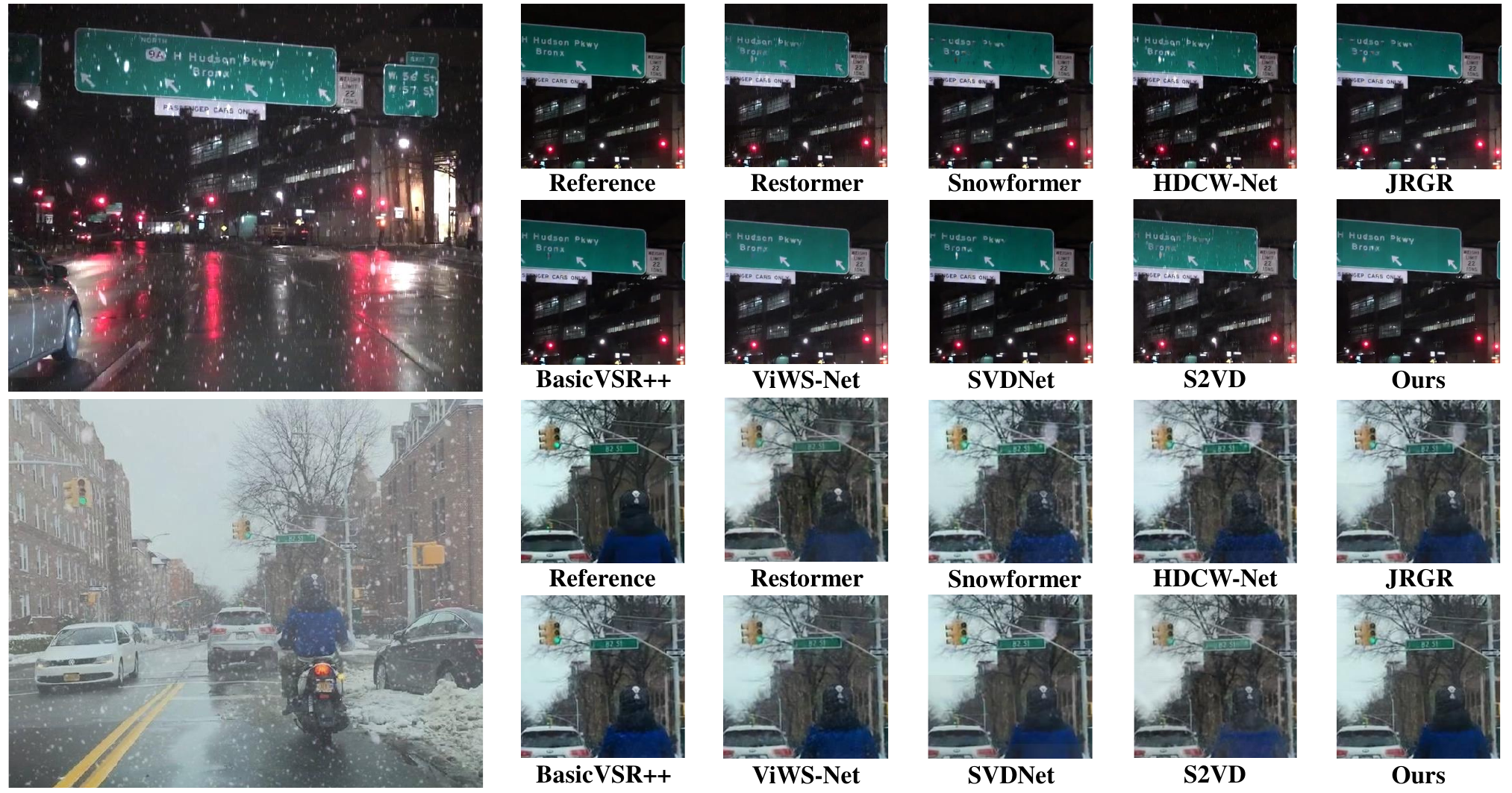}
    \caption{Visual comparisons of desnowed results produced by our network and state-of-the-art desnowing methods for input video frames from the RVSD dataset.}
    \label{fig:compare1}
    % \vspace{-3mm}
\end{figure*}

\begin{figure*}[t]
\centering
    \includegraphics[width=0.70\textwidth]{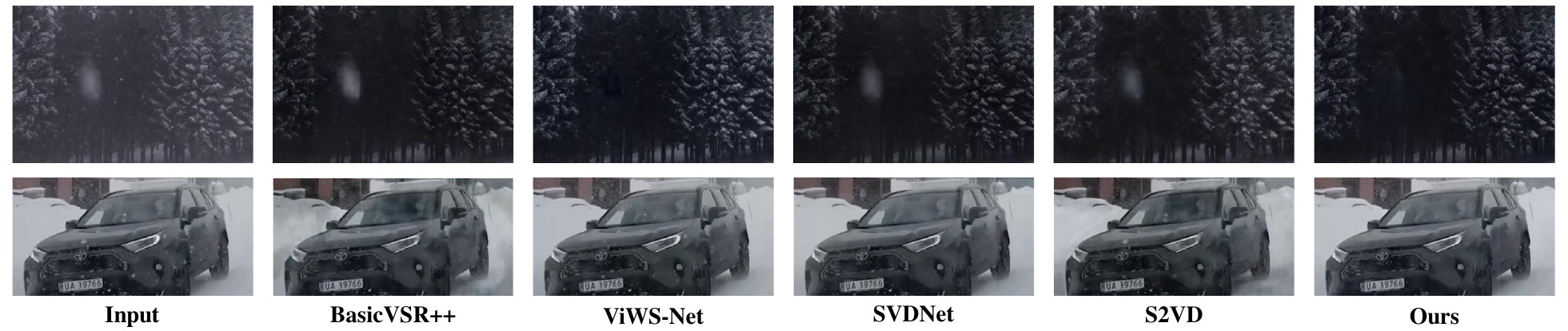}
    % \vspace{-3mm}
    \caption{ Visual comparisons of desnowed results produced by our network and state-of-the-art video desnowing methods for input video frames from real-world snowy videos.}
    \label{fig:compare2}
    % \vspace{-3mm}
\end{figure*}

\noindent{\bfseries Synthetic Datasets.} Table \ref{tab:resultcompare} reports the quantitative results of our network and $15$ state-of-the-art methods on the RVSD test dataset. 
Among these methods, SVDNet stands out as the most competitive with the highest PSNR score of 25.06 dB, the highest SSIM score of 0.9210, and the lowest LPIPS score of 0.0842.
Instead, our network outperforms SVDNet, evidenced by the higher PSNR and SSIM scores of 25.68 dB and 0.9254 respectively, as well as a lower LPIPS score of 0.0785. These results highlight the exceptional performance of our network in effectively removing snow and preserving image quality.

\noindent{\bfseries Real-World Datasets.} As shown in the comprehensive results presented in the Table \ref{tab:resultcompare}, our network also outperforms alternative methods in non-reference image quality evaluation metrics such as NIMA and MUSIQ. The comparison clearly demonstrates that our network excels in restoring images with superior quality, offering clearer content and enhanced perceptual fidelity when compared to other methods in real snowy scenarios.

\noindent{\bfseries Qualitative Comparison.} Fig.~\ref{fig:compare1} visually compares snow removal results predicted by our network and state-of-the-art methods from the RVSD dataset. Compared with other approaches, our network demonstrates superior performance in restoring the original background images by effectively eliminating snow and haze from input video frames. To further validate its efficacy on real-world data, we conduct a comparative analysis of different methods on snowy videos from our Realsnow85 testing set, as illustrated in Figs.~\ref{fig:compare2}. The results clearly indicate that our network excels in removing real snow and haze, while also successfully recovering obscured background details. Conversely, other methods tend to retain certain levels of snow and haze in their desnowed outputs.

\subsection{Ablation Study}
\noindent{\bfseries Baseline Design.} To analyze the effectiveness of our SemiVDN, we conduct ablation studies to reveal the influence of three key components in our method, i.e., the temporal decoupling experts (TDE), the semi-supervised training (SST), and the Distribution-driven Contrastive Regularization (DCR) of our SemiVDN. The first baseline network (denoted as ``M1'') is constructed by removing the temporal decoupling experts module and the teacher model, which means that only the supervised loss on labeled data is used for training.
Then, we use the temporal decoupling experts module to replace the FFN module in the transformer block to build ``M2''. After that, ``M3'' is constructed based on ``M2'' by combining semi-supervised training with the unsupervised loss in Sect. 3.3.

\begin{table}[!tbp]
\centering

\caption{Quantitative results of our network and constructed baseline networks (``M1'' to ``M3'') of the ablation study on synthetic datasets and real-world datasets.}
% \vspace{-1mm}
\label{tab:AblationStudy}
\resizebox{0.6\columnwidth}{!}{%

\begin{tabular}{c|ccc|ccc|cc} 
\hline
Method   & TDE & SST & DCR & PSNR$\uparrow$   & SSIM$\uparrow$    & LPIPS $\downarrow$  &  NIMA $\uparrow$ & MUSIQ$\uparrow$  \\ 
\hline
M1 &     &      &    & 24.41 & 0.9116 & 0.0932  &  4.165        & 49.53        \\
M2 &  \checkmark   &      &    & 25.16 & 0.9217 & 0.0822 & 4.212         & 50.69       \\
M3 &   \checkmark  & \checkmark     &    & 25.29 & 0.9237 & 0.0806 &  4.239         & 51.05       \\ 
\hline
\rowcolor{mygray}\textbf{SemiVDN}  &   \checkmark  &\checkmark      & \checkmark   & \textbf{25.68} & \textbf{0.9254} & \textbf{0.0785} &  \textbf{4.259}         & \textbf{51.57}        \\
\hline
\end{tabular}
}
% \vspace{-2mm}
\end{table}

\noindent{\bfseries Quantitative Comparison.} Table~\ref{tab:AblationStudy} reports the quantitative scores of our method and the three baseline networks (\ie, ``M1'' to ``M3''). Specifically, compared with ``M1'', ``M2'' improves the PSNR score from 24.41 dB to 25.16 dB, the SSIM score from 0.9116 to 0.9217, and the LPIPS score from 0.0932 to 0.0822. 
This demonstrates the effectiveness of the temporal decoupling experts module in decomposing the physical components of snow videos in a temporal spirit, resulting in the enhanced recovery of background.
Furthermore, our advanced "M3" model performs superior results compared to "M2", effectively showcasing the benefits of incorporating unlabeled data during the semi-supervised training to enhance the model's snow removal capabilities on synthetic and real data. 
Moreover, our network outperforms  ``M2'' and ``M3'' in terms of the six metrics, which means leveraging the three components together enables the proposed network to achieve the best performance in video snow removal on both synthetic datasets and real-world datasets.

%% file: sec/6_Conclusion.tex
\section{Conclusion}
\label{sec:intro}
In this paper, we proposed the first semi-supervised video desnowing framework named SemiVDN, which effectively leverages knowledge from unlabeled data to enhance the generalization capabilities of deep models.
To achieve superior snow removal, we incorporate the Prior-guided Temporal Decoupling Experts module, which explicitly decomposes the physical components of a snow video in a temporal manner.
Furthermore, we propose a Distribution-driven Contrastive Regularization Loss that addresses the appearance discrepancy between synthetic and real data, ensuring the preservation of snow-invariant information.
Observed from extensive experimentation on both synthesized and real-world snowy videos, our network demonstrates promising performance, surpassing existing state-of-the-art methods in snow removal.

\clearpage